\documentclass[aps,pra,twocolumn,preprintnumbers,amsmath,amssymb,footinbib,twoside]{revtex4-1}
\usepackage[utf8]{inputenc}
\usepackage{amssymb}
\usepackage[a4paper, margin=2cm, bottom=4cm,top=4cm]{geometry}
\usepackage[font=small]{caption}
\usepackage[ruled,vlined]{algorithm2e}
\usepackage{gensymb}
\usepackage{tabularx}
\usepackage{amsmath}
\usepackage{enumitem}
\usepackage{float}
\usepackage{graphicx}
\usepackage{multirow}
\usepackage{subcaption}
\usepackage{colortbl}
\usepackage{caption}
\usepackage{textcomp}
\usepackage{textgreek}
\usepackage{indentfirst}
\usepackage{listings}
\usepackage{bera}
\usepackage[table,xcdraw]{xcolor}
\usepackage[final]{hyperref}

\usepackage{fancyhdr}
\usepackage{titlesec}
\titlespacing*{\section}{5pt}{1.5\baselineskip}{0.5\baselineskip}
\titleformat*{\section}{\normalsize\bfseries\centering}

\DeclareMathOperator*{\argmax}{arg\,max}
\DeclareMathOperator*{\argmin}{arg\,min}

\hypersetup{
	colorlinks=true,
	linkcolor=blue,
	citecolor=blue,
	filecolor=magenta,
	urlcolor=blue         
}

\begin{document}

\renewcommand{\footskip}{2cm}

\setlength{\abovedisplayskip}{1.5em}
\setlength{\belowdisplayskip}{1.5em}

\arrayrulecolor{gray}

\title{
    {\Large \bf Meta-Learning Conjugate Priors for \\ Few-Shot Bayesian Optimization} \\
}

\author{
    {Ruduan B.F. Plug} \\
    \normalsize{\href{mailto:r.b.f.plug@umail.leidenuniv.nl}{r.b.f.plug@umail.leidenuniv.nl}} \\
    {Leiden Institute of Advanced Computer Science \\ Leiden University}
}

\date{03 January 2020}

\begin{abstract}
\vspace{0.5em}
\begin{center}
\begin{minipage}{\dimexpr\paperwidth-9.75cm}
\begin{center} \textbf{Abstract} \end{center}
\quad \footnotesize Bayesian Optimization is methodology used in statistical modelling that utilizes a Gaussian process prior distribution to iteratively update a posterior distribution towards the true distribution of the data. Finding unbiased informative priors to sample from is challenging and can greatly influence the outcome on the posterior distribution if only few data are available. In this paper we propose a novel approach to utilize meta-learning to automate the estimation of informative conjugate prior distributions given a distribution class. From this process we generate priors that require only few data to estimate the shape parameters of the original distribution of the data.
\vspace{0.5em}
\end{minipage}
\end{center}
\end{abstract}

\makeatletter
\@twosidetrue
\let\ps@titlepage\ps@fancy
\makeatother

\maketitle

\section{Introduction}
Meta-Learners such as those proposed by Finn. et al. \cite{Finn2017ModelAgnosticMF} and Nichol \& Schulman \cite{Nichol2018ReptileAS} provide meta-algorithms to estimate optimized initialization weights parameters of gradient-descent based learning algorithms. These optimizations are performed by ameliorating these initialization parameters on sample distributions of task classes, after which the distribution of samples from a new task of the same class can be estimated using only few data points. We can provide accurate estimates given the task is similar, as generalization of learned meta-initialization weights improve with task-similarity \cite{Khodak2019ProvableGF}.

In this paper we utilize the process of meta-learning on posterior statistical distributions of randomly generated data to estimate an informative conjugate prior. We cover numeric data under the Gaussian distributions, for which we will meta-learn a model that we hypothesize can generalize well over any parametric configuration of these distributions. Typically, it is challenging to find unbiased informative priors for Bayesian optimization tasks \cite{Thorson2017UniformUO}. We hypothesize that the estimated parameters of these meta-learned models form informative priors for these distributions \cite{Ravi2019AmortizedBM}, which has applications in the Bayes by Backpropagation technique in Bayesian Optimization \cite{Buntine1991BayesianB}.

\section{Related Work}

Previous study was done in the effects of weight initializations on the outcome of networks optimized through gradient descent, such as the early work by Drago \& Ridella on statistical methods to optimize weight initialization \cite{Drago1992StatisticallyCA}, the generalized analysis of initialization and optimization of MLPs by Weymaere \& Martens \cite{Weymaere1994OnTI} and more recently the analysis of the importance of weight initialization and learning momentum on deep learning models by Sutskever \& Martens et al.

A methodology developed for the optimization of initialization weights as a method of meta-learning was made by Finn et al. in the form of Model-Agnostic Meta Learning \cite{Finn2017ModelAgnosticMF}. Later this approach was approximated with a more efficient algorithm Reptile by Nichol \& Schulman \cite{Nichol2018ReptileAS} which has seen applications in Generative Adverserial Networks \cite{Cloutre2019FIGRFI} and Reinforcement Learning \cite{Nguyen2018ReptileFM}.

The methodology we propose in this paper builds upon the work in Amortized Bayesian Meta-Learning by Ravi \& Beatson \cite{Ravi2019AmortizedBM}, in which a method is introduced to approximate the hierarchical variation across task classes to estimate prior distributions. Other applications of meta-learning in Bayesian statistics include prior estimation for Bayesian regression by Harrison et al. \cite{Harrison2018MetaLearningPF} and \newpage using meta-learned ensemble models for Bayesian optimization methods \cite{Feurer2018ScalableMF}.

\section{Methods}

In this paper we implement a parallelized version of the scalable Reptile algorithm by Nichol \& Schulman using Torch \cite{Paszke2019PyTorchAI}. By doing so, we can process a very large amount of batch evaluations with the Reptile meta-learner while keeping computational time limited. Below in Algorithm 1 we give a general overview of the algorithm we will use to show how we update the weight initialization parameters $\theta$.

\begin{algorithm}[ht]
\SetAlgoLined
 Randomly Initialize Weight Parameters $\theta$\;
 \For{Iteration = 1, 2, ...}{
  Sample Task Batch $\tau_1, \tau_2, ... \tau_n$\;
  \For{Task = $\tau_1, \tau_2, ... \tau_n$}{
   $W_i = \text{SGD}(\mathcal{L}_{\tau_i}, \theta, k)  $\;
   }
   $\theta \leftarrow \theta + \alpha \frac{1}{k} \sum^{n}_{i=1}(W_i - \theta)$ \;
 }
 \caption{Parallel Batched Reptile {\small \cite{Nichol2018ReptileAS}}}
\end{algorithm}

\begin{figure}[ht]
    \centering
    \includegraphics[width=0.325\textwidth]{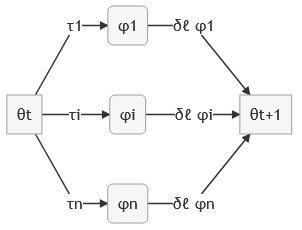}
    \caption{Schema for Batched Meta-Learning Initialization $\theta$ and Candidate $\phi$ over Tasks $\tau$}
    \label{fig:mlschema}
\end{figure}

Here tasks $\tau_1, \tau_2, ... \tau_n$ are randomly sampled task from our main task class $\mathcal{T}$. We utilize Torch to perform the SGD step, where we calculate the weights after one update over the task batch in parallel using a Torch module. From these candidate weights $\phi_i$, we calculate the linear loss gradients $\delta \ell$ by subtracting these candidate weight sets from the original weights. The mean of all the loss gradients multiplied by the learning rate gives us the final gradient for the meta-learning step resulting in parameters $\theta_{t+1}$ as illustrated in Figure \ref{fig:mlschema}.

By performing the $k$-batched optimizations, we optimize the initialization weights towards the mean of $k$ sampling distributions of the task class. As evidenced in Bronskill et al. \cite{Bronskill2020TaskNormRB}, batch updating provide a robust method even with a simple linear normalization scheme. Doing so allows us to perform meta-learning towards different task-specific goals in parallel limited to a single iteration as we are not estimating computationally expensive $n$-th order derivatives as would be done in the methodology by Finn. et al. \cite{Finn2017ModelAgnosticMF}.

\vspace{1em}

Pre-training the initialization parameters of a neural network through Meta-Learning allows for convergence using significantly less data points on MLP based models. The paper on provable guarantees for gradient-based meta-learning by Khodak et al. \cite{Khodak2019ProvableGF} further argues that generalization of the Meta-Learned parameters improve with task-similarity. We hypothesize that using this technique we can provide an automated methodology to transform these initialization weights into sampling probability distributions, which we can apply in the sampling procedure of Bayesian Optimization.

Expanding upon the work of Snoek et al. \cite{Snoek2012PracticalBO} and Shahriari et al. \cite{Shahriari2016TakingTH} we explore the possibility to generate conjugate prior distributions for the initial sampling to improve convergence using little samples, which we will consider as Few-Shot Bayesian Optimization. This approach is advantageous because it allows us to estimate the parametric configuration to fit an unknown distribution towards a chosen conjugate without requiring evaluation of sample point derivatives \cite{Mockus1989BayesianAT} as would be required in gradient-based optimization strategies that are typically used in machine learning.

\begin{algorithm}[ht]
\SetAlgoLined
 Observe Sampling Function $f(x)$ and Boundary $\mathcal{X}$\;
 \For{Iteration = 1, 2, ...}{
  Sample Prior Distribution $\mathcal{D}$\;
  Acquisition $\displaystyle x_{i+1} = \argmax_\mathcal{X} \alpha(x | \mathcal{D}_i)$\;
  Estimate Objective $y_{i+1} \sim \mathcal{N}(f(x_{i+1}),\nu)$ \;
  Augment $\mathcal{D}_{i+1} = \{\mathcal{D}_i \cup (x_{i+1},y_{i+1}) \}$ \;
 }
 \caption{Bayesian Optimisation}
\end{algorithm}

\begin{figure}[ht]
    \centering
    \includegraphics[width=0.45\textwidth]{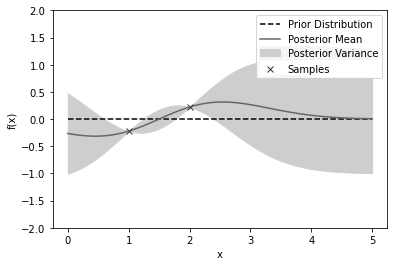}
    \caption{Bayesian Optimization from zero prior after two samples in boundary $\mathcal{X}$ = [0,5]}
    \label{fig:gprocess}
\end{figure}

In Algorithm 2 we present a general overview of the Bayesian Optimization process. Note here that the sampling function $f(x)$ provides an estimate for $y_i$ with error term as we consider the original distribution unknown. This process is illustrated in Figure \ref{fig:gprocess}, where we use an uninformative prior and have updated the Gaussian distribution with two samples.

For the Bayesian Optimizer we use the Gaussian Process Model as a non-parametric surrogate model. This model forms a mapping of boundary set $\mathcal{X}$ as $f : \mathcal{X} \to \mathbb{R}$ such that finite set $\{ x \in \mathcal{X} \}^n$ forms a $\mathbb{R}^n$ multivariate Gaussian Distribution. To guide the exploration of the sample space $\mathcal{X}$ using the sampling function, an acquisition functions $\alpha$ that is guided by the Gaussian Process mean and variance over prior $\mathcal{D}$. A typical method is to use maximum likelihood on the conjugate of the new sample and the prior distribution $\mathcal{L}(x | \mathcal{D})$ \cite{Kushner1964ANM}, which has an analytic solution for Gaussian Processes of standard normals of shape $\Phi(z)$ \cite{Snoek2012PracticalBO}:
\begin{equation}
    \alpha(x | \mathcal{D}) = \Phi(\gamma(x, \mathcal{D}))
\end{equation}
Which is distributed in $\mathbb{R}^2$ as:
\begin{equation}
    \gamma(x, \mathcal{D}) = \frac{f(\argmin_{x} f(\mathcal{D})) - \mu(\mathcal{D} \cup (x,y))}{\nu_\sigma(\mathcal{D} \cup (x,y))} 
\end{equation}
Which gives the acquisition expectation:
\begin{equation}
\begin{aligned}
    \mathbb{E}(\alpha(x | \mathcal{D})) = & \sigma(\mathcal{D} \cup (x,y))\,(\gamma(x,\mathcal{D})\Phi(\gamma(x, \mathcal{D}))) \\
    & + \mathcal{N}(\gamma,\nu)
\end{aligned}
\end{equation}

Since we wish to utilize a complex prior distribution with $\theta$ parameters as derived from our meta-learned model, using the analytic solution over $\mathbb{R}^n$ is infeasible. Instead, we will augment our estimate with the squared exponential kernel $\mathcal{K}_2$, which allows us to place strong assumptions on the co-variance of our parametric space $\theta$ while evaluating new samples $\mathbf{x}$ and estimates $\mathbf{\hat{x}}$.  
\begin{equation}
    \mathcal{K}_2(\mathbf{x}, \hat{\mathbf{x}} ) = \theta_0 \exp\left[-\frac{1}{2}\kappa(\mathbf{x}, \hat{\mathbf{x}}) \right]
\end{equation}
\begin{equation}
    \kappa(\mathbf{x}, \hat{ \mathbf{x}}) = \frac{\sum_{i=1}^{n}\left( x_i - \hat{x}_i \right)^2}{\theta_i^2}
\end{equation}
Finally, integrating allows us to evaluate monte carlo samples over the prior distribution using our meta-learned $\theta$ to find the most likely sampling candidate estimate in the parametric space for our model.
\begin{equation}
    \hat{\alpha}(x|\mathcal{D}) = \int_{}^{} \alpha(x|\theta)p(\theta|x) \, d\theta 
\label{eq:intmc}
\end{equation}

Note that now for our point acquisition using our meta-learned prior, the likelihood to draw new sample $x$ depends entirely on our $\theta$ parameters on a finite interval, which we can evaluate by direct numerical optimization \cite{Acerbi2018AnEO}.

\section{Experiment}

Our objective is to perform meta-learning on the general task structure of Gaussian distributions by generating a large amount of distributions with randomly initialized parameters $\gamma$ and $\nu$ in our boundary $\mathcal{X}$ to meta-learn a new set of parameters $\theta$. By using the acquisition estimate proposed in Equation \ref{eq:intmc} we can then transform $\theta$ into a density function over $x$ from which we can estimate the sampling likelihood distribution over our prior distribution $\mathcal{D}$ to optimize our posterior distribution in our Bayesian optimization algorithm.

Ultimately, we seek to experiment whether using this methodology can provide improvements in the error estimate of the  Bayesian optimization process when only using $k$ samples, as we defined as $k$-shot Bayesian optimization. 

As discussed previously, to facilitate our experiments we have implemented a parallelized version of the Reptile meta-learner. First, let us discuss the hyper parameters we will use for our meta-learning process, which are listed below in table \ref{tab:repparams}.

\begin{table}[ht]
\begin{tabular}{|l|l||l|l|}
\hline
\multicolumn{1}{|c|}{\textbf{Parameter}} & \multicolumn{1}{c||}{\textbf{Value}} & \multicolumn{1}{c|}{\textbf{Parameter}} & \multicolumn{1}{c|}{\textbf{Value}} \\ \hline
Inner Step Size                          & 0.02                                  & Meta Batch Size                         & 10                                    \\ \hline
Inner Batch Size                         & 5                                     & Model Size                              & 64                                    \\ \hline
Outer Step Size                          & 0.1                                   & Evaluation Iterations                   & 32                                    \\ \hline
Outer Iterations                         & 10,000                                & Evaluation Batch Size                        & 10                                \\ \hline
\end{tabular}
\caption{Torch Reptile Hyper Parameters}
\label{tab:repparams}
\end{table}

For most of the parameters, such as the model and batch sizes we follow the previous suggestions by Nichol \& Schulman \cite{Nichol2018ReptileAS}. We do however opt for a lower learning rate and a lower amount of iterations due to the task class we study being less complex than studied by Nichol, but having more variance in the parametric specifications, which means it is more beneficial for us to repeat the prior approximation as a monte carlo process as discussed in the previous section. The parameters of our task are listed in Table \ref{tab:taskparams}.

\begin{table}[ht]
\begin{tabular}{|l|l|l|l|}
\hline
\multicolumn{1}{|c|}{\textbf{Parameter}} & \multicolumn{1}{c|}{\textbf{Value}} & \multicolumn{1}{c|}{\textbf{Parameter}} & \multicolumn{1}{c|}{\textbf{Value}} \\ \hline
Task                                     & Gaussian                              & Sample Radius                           & 4                                     \\ \hline
Sample Count                             & 100,000                                 & Evaluation Range                        & [1,10]                                \\ \hline
\end{tabular}
\caption{Task Parameters}
\label{tab:taskparams}
\end{table}

Note that here the evaluation range is the range of $k$ we evaluate for our $k$-shot evaluation of our model. The sample radius defines our boundary $\mathcal{X}$ and the sample count gives our monte carlo sample size to estimate the acquisition based on our estimated model parameters $\theta$.

As the Guassian process model is a non-parametric surrogate model for our distribution, the only input we give it is our acquisition function based on our prior probability densities - estimated by our $\theta$. We will determine if there are significant differences in the performance between using these priors compared to non-informative conjugate priors such as the standard normal distribution and uniform distributions when no data has been evaluated by the model. We hypothesize that by meta-learning the prior on the task specification, we will find convergence for very low $k$ samples on the meta-learned conjugate priors.

\section{Results}

Training our meta-learner on a $1 \to 64 \to 64 \to 1$ fully connected model as proposed by Nichol et al. \cite{Nichol2018ReptileAS} gives us $4\,224$ trainable parameters. First, training the general task structure gives us probability distributions for the mean value such as illustrated below in Figure \ref{fig:mlearn1}.

\begin{figure}[ht]
    \centering
    \includegraphics[width=0.4\textwidth]{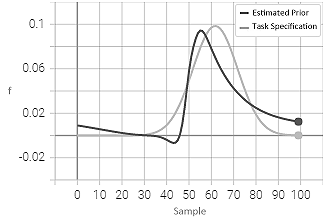}
    \caption{Estimated Prior Distribution compared to the mean of all sampled Tasks}
    \label{fig:mlearn1}
\end{figure}

These $\theta$ prior distributions on boundary $\mathcal{X}$ are based on the mean evaluation of all the batches of different task parameterizations, next we apply these to Bayesian Optimization to determine whether these generalize well over these task structures. First we evaluate Bayesian Optimization using a uniformally distributed Gaussian prior optimized using the $\mathcal{K}_2$ acquisition kernel.

\begin{figure}[ht]
    \centering
    \includegraphics[width=0.4\textwidth]{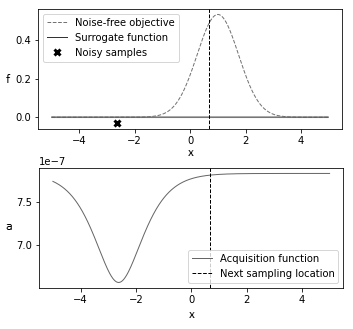}
    \caption{Initial Surrogate and Acquisition after a single sample from uniform prior distribution}
    \label{fig:bo1}
\end{figure}

\enlargethispage{10\baselineskip} 

\begin{figure}[ht]
    \centering
    \includegraphics[width=0.36\textwidth]{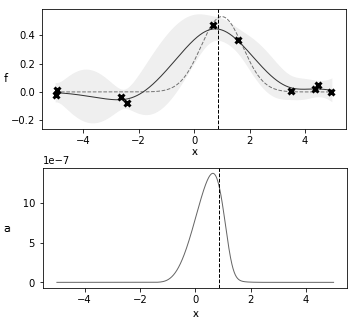}
    \caption{Final Surrogate and Acquisition after ten samples from uniform prior using $\mathcal{K}_2$ acquisition kernel}
    \label{fig:bo2}
\end{figure}

\begin{figure}[ht]
    \centering
    \includegraphics[width=0.36\textwidth]{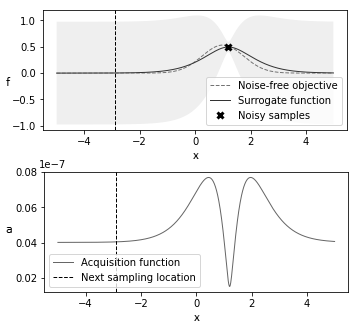}
    \caption{Initial Surrogate and Acquisition after a single sample from our meta-learned prior distribution}
    \label{fig:bo3}
\end{figure}

\begin{figure}[ht]
    \centering
    \includegraphics[width=0.36\textwidth]{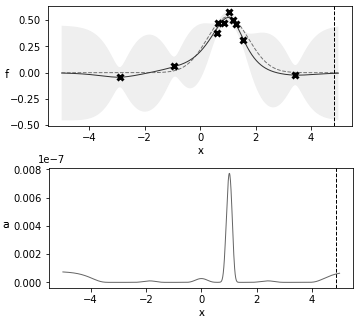}
    \caption{Final Surrogate and Acquisition after ten samples from meta prior using $\mathcal{K}_2$ acquisition kernel}
    \label{fig:bo4}
\end{figure}

In Figure \ref{fig:bo1} and \ref{fig:bo2} are shown both the Gaussian surrogate function $f$ and the acquisition $\alpha$ after $1 \cdots 10$ samples using a uniform prior. Next, we proceed with the same process using our meta-learned prior, which is shown in figures \ref{fig:bo3} and \ref{fig:bo4} and full results are provided in Table \ref{tab:res}. 
\begin{table}[ht]
\begin{tabular}{|c|c|c|c|c|c|}
\hline
\textbf{k} & \textbf{\begin{tabular}[c]{@{}c@{}}MSE\\ Uniform\end{tabular}} & \textbf{\begin{tabular}[c]{@{}c@{}}MSE\\ Meta\end{tabular}} & \textbf{k} & \textbf{\begin{tabular}[c]{@{}c@{}}MSE\\ Uniform\end{tabular}} & \textbf{\begin{tabular}[c]{@{}c@{}}MSE\\ Meta\end{tabular}} \\ \hline
1          & 1.3048                                                         & 3.1180e-04                                                  & 6          & 0.2240                                                         & 7.5502e-05                                                  \\ \hline
2          & 0.6667                                                         & 2.1578e-04                                                  & 7          & 0.2174                                                         & 5.4062e-05                                                  \\ \hline
3          & 0.4238                                                         & 1.2765e-04                                                  & 8          & 0.1597                                                         & 4.0239e-05                                                  \\ \hline
4          & 0.3412                                                         & 1.4147e-04                                                  & 9          & 0.1670                                                         & 2.9388e-05                                                  \\ \hline
5          & 0.24153                                                        & 5.8190e-05                                                  & 10         & 0.1204                                                         & 2.7168e-05                                                  \\ \hline
\end{tabular}
\caption{MSE over evaluations of Bayesian Optimization on Uniform and Meta-Learned Priors}
\label{tab:res}
\end{table}

\newpage

\section{Discussion}

From the results in the previous section we infer a noticeable difference in performance between initializing Bayesian optimization with a uniform or standard normal conjugate prior compared to the meta-learned conjugate prior. Using the meta-learned prior provides noticeably better estimates on low $k$ for $k$-shot optimization, providing better estimates with only a single sample illustrated in the results Table \ref{tab:res} and in Figure \ref{fig:bo3} than using a uniform distribution with ten samples as illustrated in Figure \ref{fig:bo2}.

\begin{figure}[ht]
    \centering
    \includegraphics[width=0.375\textwidth]{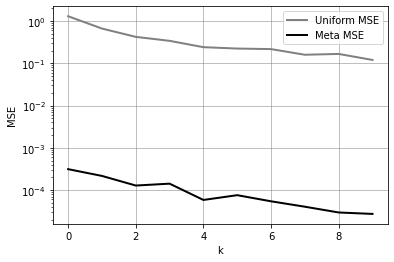}
    \caption{MSE on the evaluation set for Uniform and Meta-Learned priors show linearity with $k$ on a log scale}
    \label{fig:mse}
\end{figure}

We also note that, as expected, with a higher sample count $k$ in our evaluation the MSE over our evaluation batches decreases. This decrease in MSE is exponential with respect to $k$, which we show in the log-scale plot in Figure \ref{fig:mse}.

\section{Conclusion}

From the results we have discussed, we can indeed infer that meta-learning with reptile can provide informative conjugate priors when used for bayesian optimization. As evidenced by Table \ref{tab:res} this difference is significant when compared to the uniform baseline used in Gaussian process optimization when the task class is known. 

Concluding, meta-learning is not only a proven technique to optimize the initialization parameters for gradient-descent based methods, but the multivariate distribution mapped by the optimized parameters also provide a promising technique to map priors to multivariate gaussian distributions using the integration technique we have discussed in the methods section. Using even as little as a single sample point for a relatively simple multivariate distribution task can give us posterior distributions that are quite close to the true distribution.

\section{Further Work}

Moving forward, we may be able to experiment to use this technique towards non-conjugate priors and non-standard distributions. The most likely candidate is a multivariate beta distribution, which is a viable Bayesian prior for a large variety of tasks such as image processing proposed by Akhtar et al. \cite{Akhtar2016HierarchicalBP}, sparse learning as shown by Yuan et al. \cite{Yuan2013MultitaskSL} and multistate events covered by Kim et al. \cite{Kim2012BayesianAO}. 

Comparing different meta-learning techniques for optimizing the priors in Bayesian optimization or varying the prior surrogate from GP may provide novel results as illustrated by Wu \cite{Wu2020ALPaCAVG} comparing Bayesian meta-learning algorithms.

\bibliographystyle{ieeetr}

\bibliography{bib}

\end{document}